\pdfoutput=1

\documentclass[11pt]{article}

\usepackage[]{acl}

\usepackage{times}
\usepackage{latexsym}
\usepackage{graphicx}
\usepackage{multirow}
\usepackage{hyperref}
\usepackage{booktabs}

\usepackage[T1]{fontenc}

\usepackage[utf8]{inputenc}

\usepackage{microtype}

\title{Average Is Not Enough: Caveats of Multilingual Evaluation}

\author{Matúš Pikuliak \and Marián Šimko \\ Kempelen Institute of Intelligent Technologies \\ \texttt{firstname.surname@kinit.sk}}

\begin{document}
\maketitle
\begin{abstract}
This position paper discusses the problem of multilingual evaluation. Using simple statistics, such as average language performance, might inject linguistic biases in favor of dominant language families into evaluation methodology. We argue that a qualitative analysis informed by comparative linguistics is needed for multilingual results to detect this kind of bias. We show in our case study that results in published works can indeed be linguistically biased and we demonstrate that visualization based on URIEL typological database can detect it.
\end{abstract}

\section{Introduction}

\begin{figure*}[t]
\includegraphics[width=16cm]{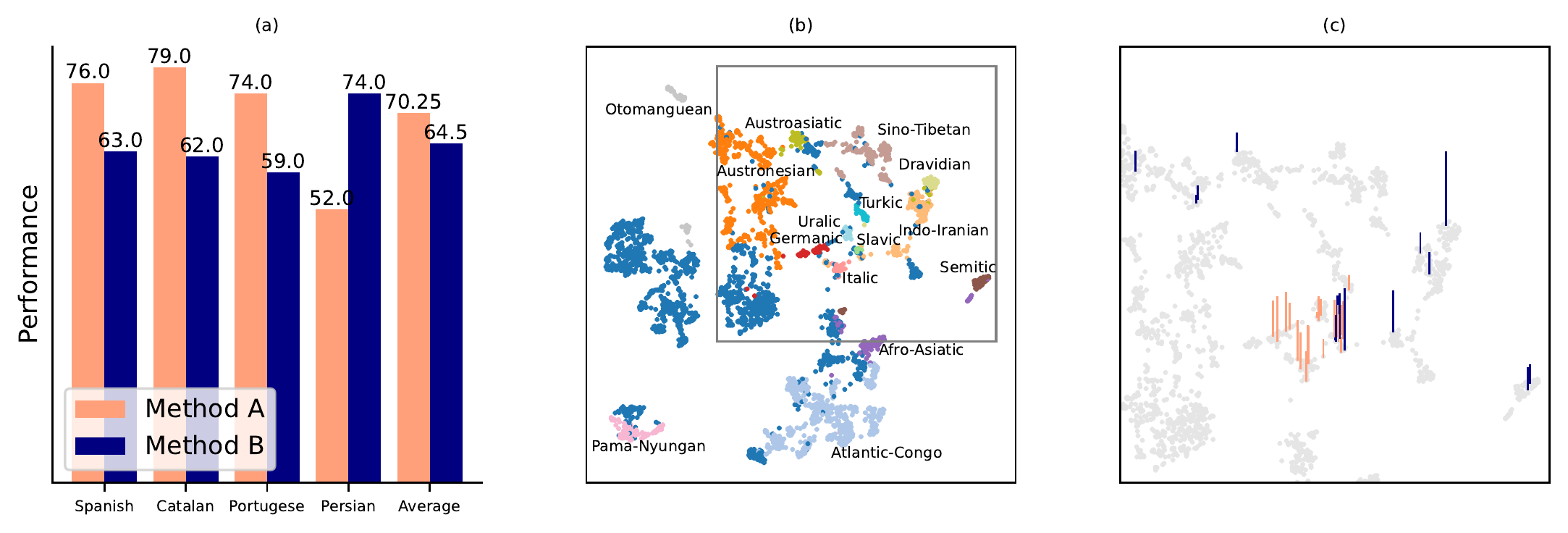}
\caption{\textit{(a)} Comparison of two methods on unbalanced set of languages. \textit{(b)}  Visualization of URIEL languages with certain language families color-coded. \textit{(c)} Comparison of two methods from~\citeauthor{rahimi} This uses the same map of languages as \textit{b}, but the view is zoomed.}\label{fig2}
\end{figure*}

The linguistic diversity of NLP research is growing~\cite{joshi-etal-2020-state,PIKULIAK2021113765} thanks to improvements of various multilingual technologies, such as machine translation~\cite{DBLP:journals/corr/abs-1907-05019}, multilingual language models~\cite{devlin-etal-2019-bert, DBLP:conf/nips/ConneauL19}, cross-lingual transfer learning~\cite{PIKULIAK2021113765} or language independent representations~\cite{DBLP:journals/jair/RuderVS19}. It is now possible to create well-performing multilingual methods for many tasks. When dealing with multilingual methods, we need to be able to evaluate how good they really are, i.e. how effective they are on a wide variety of typologically diverse languages. Consider the two methods shown in Figure~\hyperref[fig2]{1 (a)}. Without looking at the particular languages, \textit{Method A} seems better. It has better results for the majority of languages and its average performance is better as well. However, the trio of languages, where \textit{Method A} is better, are in fact all very similar Iberian languages, while the fourth language is Indo-Iranian. Is the \textit{Method A} actually better, or is it better only for Iberian? Simple average is often used in practice without considering the linguistic diversity of the underlying selection of languages, despite the fact that many corpora and datasets are biased in favor of historically dominant languages and language families.

Additionally, as the number of languages increases, it is harder and harder to notice phenomena such as this. Consider the comparison of two sets of results in Table~\ref{tab:results}. With 41 languages it is cognitively hard to discover various relations between the languages and their results, even if one has the necessary linguistic knowledge.


\begin{table*}[]
\setlength{\tabcolsep}{3pt}
\tiny
    \centering
    \begin{tabular}{r|ccccccccccccccccccccc}
    \toprule
Language & afr & arb & bul & ben & bos & cat & ces & dan & deu & ell & eng & spa & est & pes & fin & fra & heb & hin & hrv & hun & ind \\
Method A & 74 & 54 & 54 & 60 & 77 & 79 & 72 & 79 & 64 & 34 & 57 & 76 & 71 & 52 & 69 & 73 & 46 & 58 & 77 & 69 & 61 \\
Method B & 59 & 64 & 61 & 70 & 63 & 62 & 62 & 62 & 58 & 61 & 47 & 63 & 64 & 74 & 67 & 57 & 53 & 68 & 61 & 59 & 67  \\ \midrule
Language & ita & lit & lav & mkd & zlm & nld & nor & pol & por & ron & rus & slk & slv & alb & swe & tam & tgl & tur & ukr & vie & \textbf{AVG} \\
Method A & 76 & 75 & 67 & 48 & 63 & 78 & 77 & 77 & 74 & 74 & 36 & 76 & 76 & 76 & 69 & 25 & 57 & 67 & 49 & 48 & 64.5 \\
Method B & 60 & 62 & 68 & 67 & 66 & 59 & 65 & 61 & 59 & 66 & 53 & 62 & 64 & 69 & 69 & 54 & 66 & 61 & 60 & 55 & 62.1 \\
\bottomrule
    \end{tabular}
    \caption{Comparison of two methods from~\citet{rahimi}.}
    \label{tab:results}
\end{table*}

In this position paper, we argue that it is not the best practice to compare multilingual methods only with simple statistics, such as average. Commonly used simple evaluation protocols might bias research in favor of dominant languages and in turn hurt historically marginalized languages. Instead, we propose to consider using qualitative results analysis that takes linguistic typology~\cite{ponti-etal-2019-modeling} and comparative linguistics into account as an additional sanity check. We believe that this is an often overlooked tool in our research toolkit that should be used more to ensure that we are able to properly interpret results from multilingual evaluation and detect various linguistic biases and problems. In addition to this discussion, which we consider a contribution in itself, we also propose a visualization based on URIEL typological database~\cite{littell-etal-2017-uriel} as an example of such qualitative analysis, and we show that it is able to discover linguistic biases in published results.

\section{Related Work}

\paragraph{Linguistic biases in NLP.} \citet{bender-2009-linguistically} postulated that research driven mainly by evaluation in English will become biased in favor of this language and it might not be particularly language independent. Even in recent years, popular techniques such as \textit{word2vec} or \textit{Byte Pair Encoding} were shown to have worse performance on morphologically rich languages~\cite{bojanowski-etal-2017-enriching, park2020morphology}. Similarly, cross-lingual word embeddings are usually constructed with English as a default hub language, even though this might hurt many languages~\cite{anastasopoulos-neubig-2020-cross}. Perhaps if the practice of research was less Anglocentric, different methods and techniques would have become popular instead. Our work is deeply related to issues like these. We show that multilingual evaluation with an unbalanced selection of languages might cause similar symptoms.

\paragraph{Benchmarking.} Using benchmarks is a practice that came under a lot of scrutiny in the NLP community recently. Benchmark evaluation was said to encourage spurious data overfitting~\cite{kavumba-etal-2019-choosing}, encourage metric gaming~\cite{DBLP:journals/corr/abs-2002-08512} or lead the research away from general human-like linguistic intelligence~\cite{linzen-2020-accelerate}. Similarly, benchmarks are criticized for being predominantly focused on performance, while neglecting several other important properties, e.g. prediction cost or model robustness~\cite{ethayarajh-jurafsky-2020-utility}. Average in particular was shown to have several issues with robustness that can be addressed by using pair-wise instance evaluation~\cite{peyrard}. To address these issues, some benchmarks refuse to use aggregating scores and instead report multiple metrics at the same time leaving interpretation of the results to the reader. \citet{gem} is one such benchmark, which proposes to use visualizations to help the intepretation. In this work, we also use visualizations to similar effect.

\section{Multilingual Evaluation Strategies}

When comparing multilingual methods with non-trivial number of languages, it is cognitively hard to keep track of various linguistic aspects, such as language families, writing systems, typological properties, etc. Researchers often use various simplifying strategies instead:

\paragraph{Aggregating metrics.} Aggregating metrics, such as average performance or a number of languages where a certain method achieves the best results provide some information, but as we illustrated in Figure~\hyperref[fig2]{1 (a)}, they might not tell the whole story. By aggregating results we lose important information about individual languages and language families. Commonly used statistics usually do not take underlying linguistic diversity into account. This might lead to unwanted phenomena, such as bias in favor of dominant language families. The encoded values of the aggregating metrics might not align with the values we want to express. Average is an example of utilitarianist world view, while using minimal performance might be considered to be a prioritarianist approach~\cite{mlms}. Even though analyzing the values encoded in metrics is a step towards a fairer evaluation, they still miss a more fine-grained details of the results.

\paragraph{Aggregated metrics for different groups.} Another option is to calculate statistics for certain linguistic families or groups. These are steps in the right direction, as they provide a more fine-grained picture, but there are still issues left. It is not clear which families should be selected, e.g. should we average all Indo-European languages or should we average across subfamilies, such as Slavic or Germanic. This selection is ultimately opinionated and different selections might show us different views of the results. In addition, aggregating across families might still hide variance within these families. Grouping languages by the size of available datasets (e.g. low resource vs. high resource) shows us how the models deal with data scarcity, but the groups might still be linguistically unbalanced.

\paragraph{Balanced language sampling.} Another option is to construct a multilingual dataset so that it is linguistically balanced. This process is called \textit{language sampling}~\cite{rijkhoff1993method, miestamo2016sampling}. In practice, this means that a small number of representative languages is selected for each family. The problem with dominant families is solved because we control the number of languages per family. However, selecting which families should be represented and then selecting languages within these families is again an opinionated process. Different families and their subfamilies might have different degrees of diversity. Different selections might favor different linguistic properties and results might vary between them. It is also not clear, how exhaustive given selection is, i.e. how much of the linguistic variety has been covered. Some of the existing works mention their selection criteria:~\citet{longpre2020mkqa} count how many speakers the selection covers,~\citet{clark-etal-2020-tydi} use a set of selected typological properties,~\citet{ponti-etal-2020-xcopa} use the so called \textit{variety language sampling}. Publishing the criteria allows us to do a post-hoc analysis in the future to evaluate, how well did these criteria work.


\paragraph{Qualitative analysis} In this paper, we argue that qualitative analysis is an often overlooked, yet irreplaceable evaluation technique. In the following section, we will present our case study of how to perform qualitative analysis.

\section{Case Study: Qualitative Analysis through Visualization}\label{sec:qual}

In this section we show how to perform a qualitative analysis of multilingual results with a visualization technique based on URIEL typographic database. We show that using this we can (1) uncover linguistic biases in the results, and (2) make sense of results from non-trivial number of languages. As case study, we study results from ~\citet{rahimi}. Our goal is not to evaluate particular methods from this paper, but to demonstrate how linguistically-informed analysis might help researchers gain insights into their results. We analyze the results from this paper not because we want to criticize it, but because it is a well-written paper that actually attempts to do multilingual evaluation for non-trivial number of languages with significantly different methods. The linguistic biases we uncover are already partially discussed in the paper. Here, we only show how to effectively perform qualitative analysis and uncover these biases with appropriate visualization. Appendix~\ref{app:papers} shows similar analysis for another paper~\cite{heinzer} where linguistic biases are visible.

We use URIEL, a typological language database that consists of 289 syntactic and phonological binary features for 3718 languages. We use UMAP feature reduction algorithm~\cite{DBLP:journals/corr/abs-1802-03426} to create a 2D typological language space. This map is shown in Figure~\hyperref[fig2]{1 (b)}. The map is interactive and allows for dynamic filtering of languages and families, as well as inspection of individual languages and their properties.\footnote{\href{https://github.com/matus-pikuliak/multilingual-evaluation}{Code available at GitHub}} Each point is one language and selected language families are color-coded in the figure. Even though URIEL features used for dimensionality reduction do not contain information about language families, genealogically close languages naturally form clusters in our visualization. Certain geographical relations are captured as well, e.g. Sudanic and Chadic languages are neighboring clusters, despite being from different language families. This evokes the linguistic tradition of grouping languages according to the regions and macroregions. This shows that our visualization is able to capture both intrafamiliar and interfamiliar similarities of languages and is thus appropriate for our use-case.

We visualize results from~\citet{rahimi} on this linguistic map. \citeauthor{rahimi} use Wikipedia-based corpus for NER, and they compare various cross-lingual transfer learning algorithms for 41 languages. They use an unbalanced set of languages, where the three most dominant language families -- Germanic, Italic and Slavic -- make up 55\% of all languages. See Appendix~\ref{app:papers} for more details about the paper. We use our URIEL map to visualize a comparison between a pair of methods on all 41 languages from Table~\ref{tab:results}. In Figure~\hyperref[fig2]{1 (c)} we compare two methods -- \textit{Method A} -- cross-lingual transfer learning methods using multiple source languages (average performance $64.5$), and seemingly worse \textit{Method B} -- a low-resource training without any form of cross-lingual supervision (average performance $62.1$). We use the same URIEL map, but we superimpose the relative performance of the two methods as colored columns. Orange columns on this map show languages where \textit{Method A} performs better, while blue columns show the same for \textit{Method B}. Height of each column shows how big the relative difference in performance is between the two methods. I.e. taller orange columns mean dominant \textit{A}, taller blue columns mean dominant \textit{B}.

We can now clearly see that there is a pattern in the location of the colored columns. Using average as evaluation measure, \textit{Method A} seems better overall. Here we can see that it is only better in one particular cluster of languages -- the cluster of orange columns. All these are related European languages. Most of them are Germanic, Italic or Slavic, with some exceptions being languages that are not Indo-European, but are nevertheless geographical neighbors, such as Hungarian. On the other hand, all the non-European languages actually prefer \textit{Method B}. These are the blue columns scattered in the rest of the space that consists of languages such as Arabic (Semitic), Chinese (Sino-Tibetan) or Tamil (Dravidian).

This shows important fact about the two methods that was hidden by using average. Cross-lingual supervision seemed to have better performance, but it has better performance only in the dominant cluster of similar languages where the cross-lingual supervision is more viable. Other languages, would actually prefer using monolingual low-resource learning, as they are not able to learn from other languages that easily. In this case, average is overestimating the value of cross-lingual learning for non-European languages. This overestimation might cause harm to these languages.

We can also see that there are some exceptions -- the blue columns in the orange cluster. These exceptions are Greek, Russian, Macedonian, Bulgarian and Ukrainian -- all Indo-European languages that use non-Latin scripts. In this case, different writing systems are probably cause of additional linguistic bias. It might be hard to notice this pattern by simply looking at the table of results, but here we can quickly identify the languages as outliers and then it is easy to realize what they have in common. 

Note that we do not expect to see this level of linguistic bias in most papers and we have cherry-picked this particular methods from this particular paper because they demonstrate the case when the linguistic bias in the results is the most obvious. This is caused mainly by unbalanced selection of languages on Wikipedia and in a sense unfair comparison of cross-lingual supervision with low resource learning.





\section{Conclusions}

Multilinguality in NLP is becoming more common and methodological practice is sometimes lagging behind~\cite{artetxe-etal-2020-call, keung-etal-2020-dont, bender2011achieving}. Making progress will be inherently hard without proper evaluation methodology. In this work, we argue for necessity for qualitative results analysis and we showed how to use such analysis to improve the evaluation with interactive visualizations. In our case study, we were able to uncover linguistic biases in published results.

Considering the practice in machine learning and NLP, it might be tempting to reduce a multilingual method performance to a single number. However, we believe that intricacies of multilingual evaluation can not be reduced so easily. There are too many different dimensions that need to be taken into consideration and NLP researchers should understand these dimensions. We believe that appropriate level of training in various linguistic fields, such as typology or comparative linguistics, is necessary for proper understanding of multilingual results and for proper qualitative analysis. We argue that qualitative analysis is an oft overlooked approach to results analysis that should be utilized more to prevent various distortions in how we understand linguistic implications of our results.


\section{Ethical Considerations}

Much of current NLP research is focused on only a small handful of languages. Communities of some language users are left behind, as a result of data scarcity. We believe that our paper might have positive societal impact. It focuses on the issues of these marginalized languages and communities. Following our recommendations might lead to a more diverse and fair multilingual evaluation both in research and in industry. This might in turn led to better models, applications and ultimately quality of life changes for some.





\section*{Acknowledgments}

This research was partially supported by DisAi, a~project funded by Horizon Europe under \href{https://doi.org/10.3030/101079164}{GA No. 101079164}.


\bibliographystyle{acl_natbib}
\bibliography{custom}

\appendix
\newpage
\section{Details of Analysed Papers}\label{app:papers}

In this appendix, we provide additional information about papers we analysed.

\subsection{Rahimi et al.}

\begin{figure}[t]
\includegraphics[width=8cm]{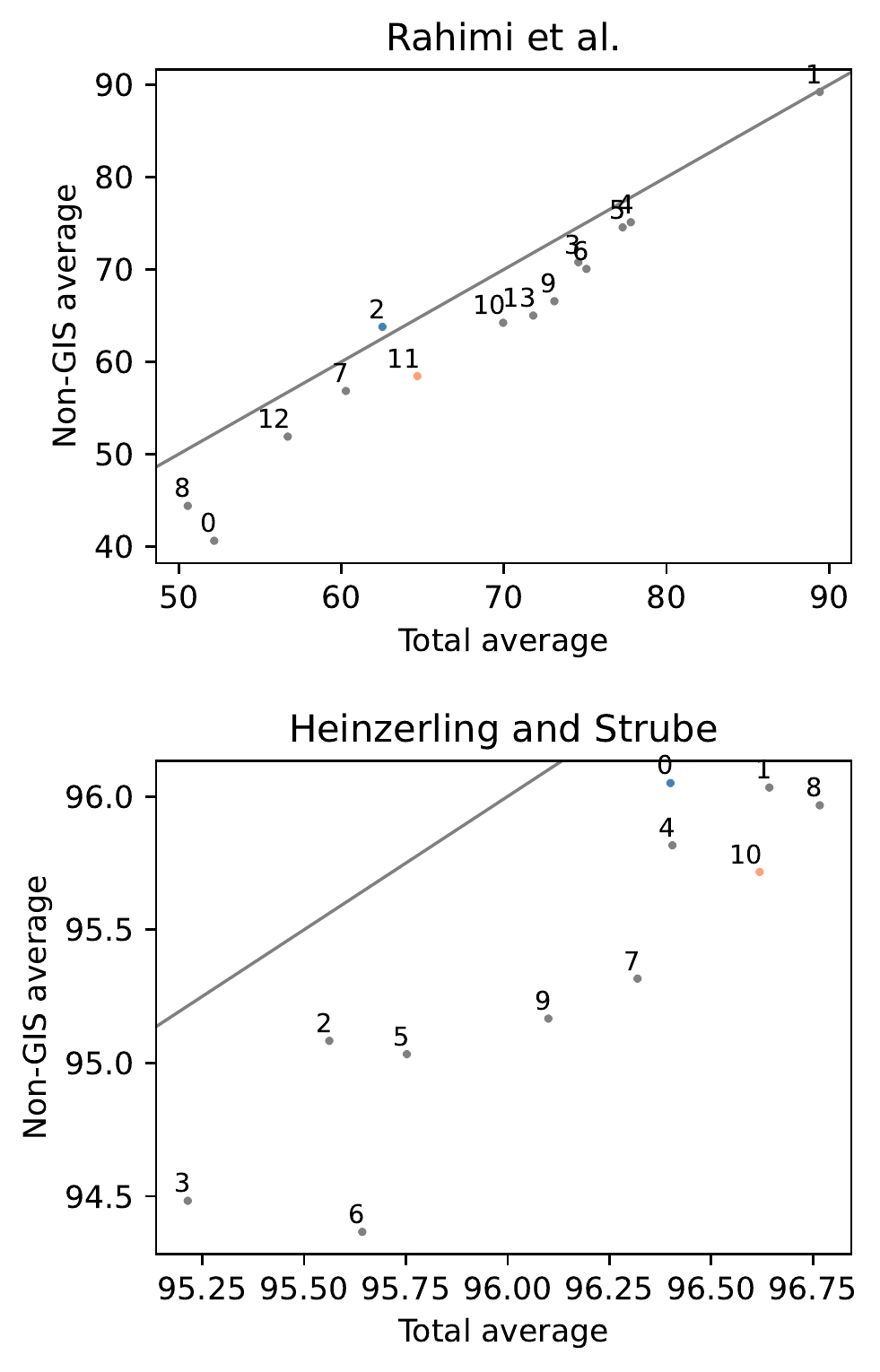}
\caption{Comparison of method performance. The relation between global average and average on non-GIS languages is shown. Each point represents one method from the papers.}\label{fig3}
\end{figure}

This is the paper we used for demonstration in the main paper in Section~\ref{sec:qual}. We use results reported in Table 4 in their paper. The languages they use are listed here in Table~\ref{tab:rahimi}. We can see the apparent dominance of Indo-European languages. There are 14 different methods listed in their paper. We compare the results for these methods in Figure~\ref{fig3}. There we can see how the average results for individual methods compare with the average results for non-GIS (Germanic-Italic-Slavic) languages. The numbers correspond to the order of methods listed in the original paper. The two methods compared in Figure~\hyperref[fig2]{1 (c)} are shown as blue and orange, respectively. The orange \textit{Method A} is \texttt{BEA\textsuperscript{tok}} in the original paper. The blue \textit{Method B} is called \texttt{LSup}. We can see the linguistic bias with this simplistic view as well. All the cross-lingual learning based methods have worse non-GIS results than methods that do not use cross-lingual learning (methods 1 and 2). However, this analysis can not replace the visualization we propose in Section~\ref{sec:qual}. It provides a GIS-centered view, but it can not capture other sources of bias. For example, it does not show various outliers that were seen in the visualization, such as Uralic languages that behave similarly to GIS languages, or Slavic languages with Cyrilic alphabet that behave differently than other Slavic languages.

\begin{table}
\small
\begin{tabular}{|l|l|l|l|}
\hline
\textbf{ISO} & \textbf{Language} & \textbf{Subfamily} & \textbf{Family} \\ \hline
bul & Bulgarian & \multirow{10}{*}{Slavic} & \multirow{28}{*}{Indo-European} \\
bos & Bosnian & & \\
ces & Czech & & \\
hrv & Croatian & & \\
mkd & Macedonian & & \\
pol & Polish & & \\
rus & Russian & & \\
slk & Slovak & & \\
slv & Slovenian & & \\
ukr & Ukrainian & & \\ \cline{1-3}
afr & Afrikkans & \multirow{6}{*}{Germanic} & \\
dan & Danish & & \\
deu & German & & \\
nld & Dutch & & \\
nor & Norwegian & & \\
swe & Swedish & &  \\ \cline{1-3}
cat & Catalan & \multirow{6}{*}{Italic} &  \\
fra & French & &  \\
ita & Italian & &  \\
por & Portugese & & \\
rom & Romanina & &  \\
spa & Spanish & &  \\ \cline{1-3}
ben & Bengali & \multirow{3}{*}{Indo-Iranian} &  \\
hin & Hindi & & \\
pes & Iranian Persian & & \\ \cline{1-3}
lit & Lithuanian & \multirow{2}{*}{Baltic} &  \\
lav & Latvian & & \\ \cline{1-3}
ell & Greek & & \\ \cline{1-3}
alb & Albanian & & \\ \hline
est & Estonian & & \multirow{3}{*}{Uralic} \\
fin & Finnish & & \\  
hun & Hungarian & & \\ \hline
ind & Indonesian & & \multirow{3}{*}{Austronesian} \\
tgl & Tagalog & & \\
zlm & Malay & & \\ \hline
arb & Standard Arabic & & \multirow{2}{*}{Afro-Asiatic} \\
heb & Hebrew & & \\ \hline
vie & Vietnamese & & Austroasiatic \\ \hline
tam & Tamil & & Davidian \\ \hline
tur & Turkish & & Turkic \\ \hline
\end{tabular}
\caption{Languages used in~\citeauthor{rahimi}.}\label{tab:rahimi}
\end{table}

\begin{table}
\small
\begin{tabular}{|l|l|l|l|}
\hline
\textbf{ISO} & \textbf{Language} & \textbf{Subfamily} & \textbf{Family} \\ \hline
dan & Danish & \multirow{6}{*}{Germanic} & \multirow{17}{*}{Indo-European} \\
deu & German &  & \\
eng & English &  & \\
nld & Dutch &  & \\
nor & Norwegian &  & \\
swe & Swedish &  &  \\ \cline{1-3}
bul & Bulgarian & \multirow{5}{*}{Slavic} & \\
ces & Czech &  & \\
hrv & Croatian &  & \\
pol & Polish &  & \\
slv & Slovenian &  & \\ \cline{1-3}
fra & Frech & \multirow{4}{*}{Italic} & \\
ita & Italian &  & \\
por & Portugese &  & \\
spa & Spanish &  & \\ \cline{1-3}
hin & Hindi & \multirow{2}{*}{Indo-Iranian} & \\
pes & Iranian Persian &  &  \\ \hline
eus & Basque & & Isolate \\ \hline
fin & Finnish & & Uralic \\ \hline
heb & Hebrew & & Afro-Asiatic \\ \hline
ind & Indonesian & & Austronesian \\ \hline
\end{tabular}
\caption{Languages used in~\citeauthor{heinzer}.}\label{tab:heinzer}
\end{table}

\subsection{Heinzerling and Strube}

\begin{figure}[t]
\centering
\includegraphics[width=7cm]{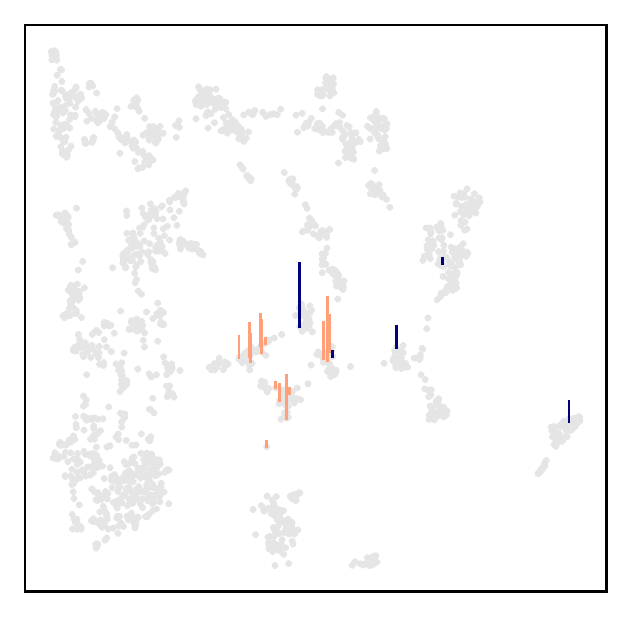}
\caption{Comparison of two methods from~\citeauthor{heinzer}.}\label{fig4}
\end{figure}

Similar linguistic biases can be seen in~\citeauthor{heinzer} as well. They evaluate various representations performance on POS tagging and NER. In Figure~\ref{fig4} we compare POS accuracy of a multilingual model with a shared embedding vocabulary (average performance $96.6$, \texttt{MultiBPEmb +char +finetune} in the original paper) and a simple BiLSTM baseline with no transfer supervision (average performance $96.4$, \texttt{BiLSTM} in the original paper). Orange columns are for languages that prefer the multilingual model, blue columns prefer the baseline. In this case, almost all orange columns are in fact GIS languages. Other languages are having significantly worse results with this method and most of them actually prefer the simple baseline with no cross-lingual supervision. This shows the limitations of proposed multilingual supervision for outlier languages.

We use results reported in Table 5 in their paper. The languages they use are listed here in Table~\ref{tab:heinzer}. Again, we can see an apparent dominance of GIS languages. There are 11 different methods listed in their paper. We omitted results for additional 6 low resource languages reported in Table 7, because only 4 out of 11 methods were used there. We compare the results for these methods in Figure~\ref{fig3}, similarly as in the previous paper. The orange point is the multilingual model, the blue point is the baseline. Now we can see that the BiLSTM baseline is actually the best performing method for non-GIS languages.

\section{Hyperparameters}

We use UMAP python library\footnote{\url{umap-learn.readthedocs.io}} with the following hyperparameters:

\begin{itemize}
    \item Number of neighbours (\texttt{n\_neighbors}): 15
    \item Distance metric (\texttt{metric}): cosine 
    \item Minimal distance (\texttt{min\_dist}): 0.5
    \item Random see (\texttt{random\_state}): 1
\end{itemize}

\section{Additional Visualizations}\label{app:vis}

In this Section we show several additional possibilities of using URIEL map of languages to visualize results from multilingual evaluation. Our goal here is to propose additional techniques that can be used for qualitative analysis apart from the comparison of two methods used in Figure~\ref{fig2} in the main body of this paper. This is not an exhaustive list of visualizations. We believe that many other types of visualization can be done using this type of qualitative analysis, based on the needs and requirements of the user.

\begin{figure*}[t]
\includegraphics[width=16cm]{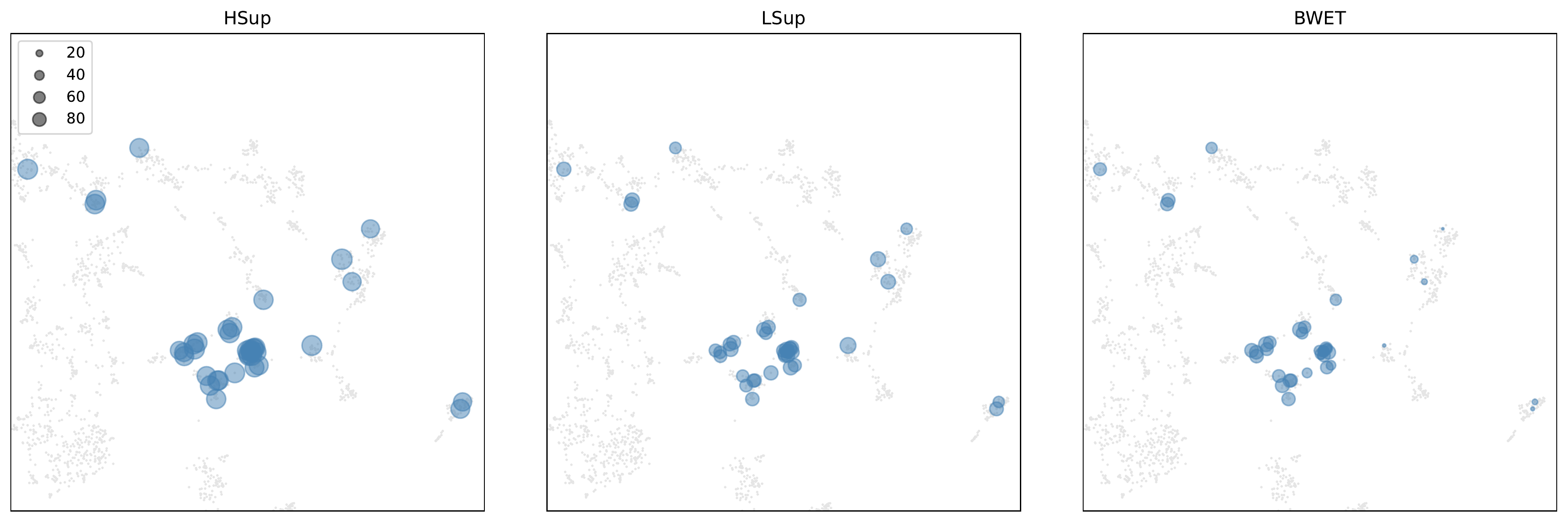}
\caption{Comparison of multiple methods using size to mark method performance for individual languages. \texttt{HSup}, \texttt{LSup} and \texttt{BWET} are methods reported in~\citep{rahimi}.}\label{fig5}
\end{figure*}

In Figure~\ref{fig5} we show how to compare more than two methods by visualizing the performance for each method separately. We have created a separate plot for three methods and we can compare their performance visually. We can see that \texttt{HSup} method has overall stable high performance. \texttt{LSup} has worse performance, but its still quite balanced. Finally, \texttt{BWET} has similar performance as \texttt{LSup}, but we can see that there are regions where it fails, e.g. the languages in the rightmost part of the figure have visibly worse performance.

\begin{figure}[t]
\includegraphics[width=7cm]{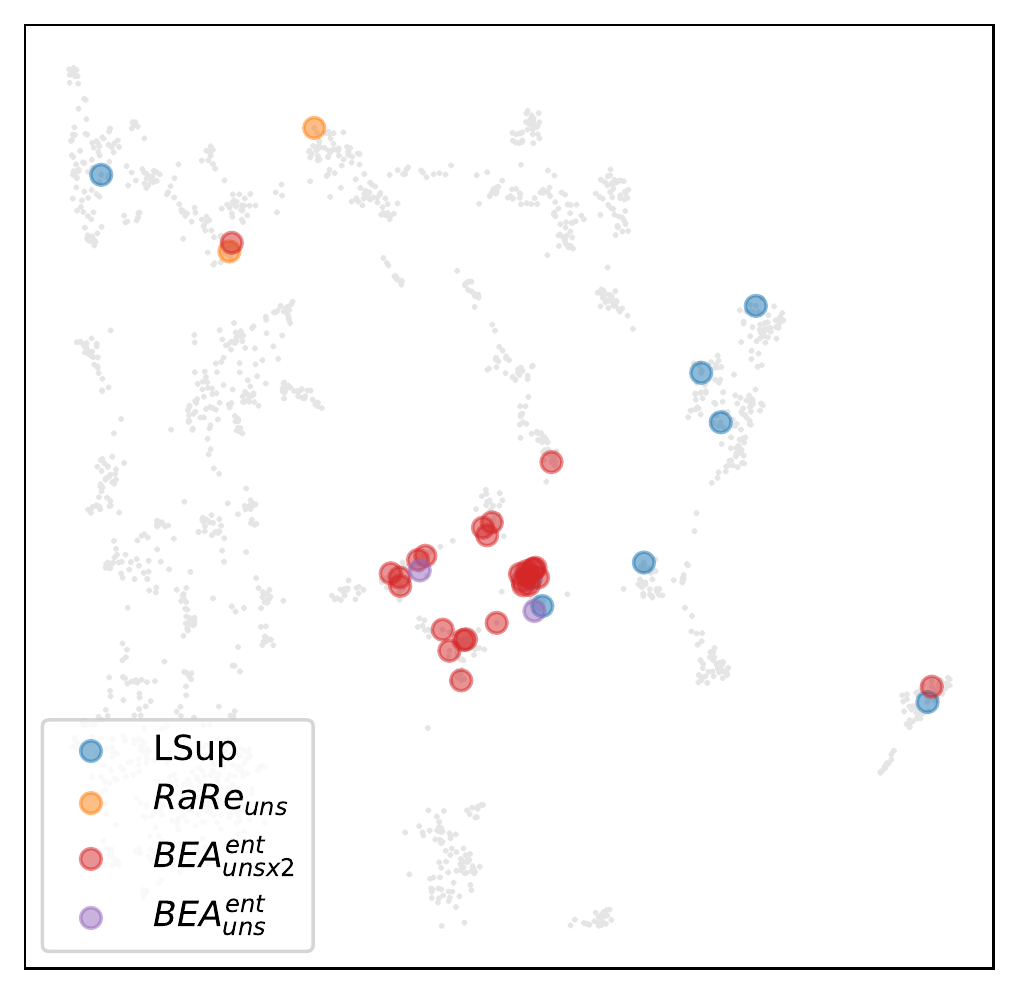}
\caption{The best performing methods for various languages.}\label{fig6}
\end{figure}

In Figure~\ref{fig6} we show yet another type of visualization. In this case, we simply visualize what method is the best performing for each language. We compare methods using crosslingual supervision and low-resource training (\texttt{LSup}). From seven methods, only four achieved the best performance for at least one language and those are shown in the Figure. Again, we can see similar picture as before. One method ($BEA^{ent}_{uns \times 2}$) is the best performing method taking average into account. However, in this visualization we can see that it is actually the best performing method only in the dominant cluster of European languages. Elsewhere, other methods perform better.

\end{document}